# Improving Medical Short Text Classification with Semantic Expansion Using Word-Cluster Embedding

Ying Shen[1], Qiang Zhang[1], Jin Zhang[1], Jiyue Huang[1], Yuming Lu[2(✉)] and Kai Lei[1]

[1] Institute of Big Data Technologies Shenzhen Key Lab for Cloud Computing Technology & Applications School of Electronics and Computer Engineering (SECE)
Peking University, SHENZHEN 518055 P.R.CHINA
`{shenying,leik}@pkusz.edu.cn`
`{zhangqiang, 1701213660, 1701213596}@sz.pku.edu.cn`
[2] ShenZhen Key Lab for Visual Media Processing and Streaming Media, ShenZhen Institute of Information Technology, SHENZHEN 518172 P.R.CHINA
`luyuming@sziit.edu.cn`

**Abstract.** Automatic text classification (TC) research can be used for real-world problems such as the classification of in-patient discharge summaries and medical text reports, which is beneficial to make medical documents more understandable to doctors. However, in electronic medical records (EMR), the texts containing sentences are shorter than that in general domain, which leads to the lack of semantic features and the ambiguity of semantic. To tackle this challenge, we propose to add word-cluster embedding to deep neural network for improving short text classification. Concretely, we first use hierarchical agglomerative clustering to cluster the word vectors in the semantic space. Then we calculate the cluster center vector which represents the implicit topic information of words in the cluster. Finally, we expand word vector with cluster center vector, and implement classifiers using CNN and LSTM respectively. To evaluate the performance of our proposed method, we conduct experiments on public data sets TREC and the medical short sentences data sets which is constructed and released by us. The experimental results demonstrate that our proposed method outperforms state-of-the-art baselines in short sentence classification on both medical domain and general domain.

## 1 Introduction

Short text classification has been proven promising in natural language processing tasks, such as social network sentiment analysis [1], product review classification [2] etc. However, the medical short texts usually contain synonym, alias and acronym but lack contextual information, leading to semantic ambiguity.

Moreover, there are many oral expressions or imprecise descriptions in the medical field. For example, acquired immune deficiency syndrome is usually abbreviated as AIDS. According to the CBOW model, the vector of the target word is predicted based on context vectors. These words have semantic similarity and often appear in the same context. Therefore, they have similar word vectors and are close in semantic space. In the semantic space, similar words are more likely to be grouped together and different words are more likely to be aggregated into different clusters.

To alleviate these problems, we propose the cluster-based word vector expansion method, which uses Hierarchical Agglomerative Clustering (HAC) to cluster the words by calculating the distance between words in the semantic space. Concretely, it calculates the cluster center vector as the potential theme information of words in this cluster. Then it adds cluster center vector to the corresponding word vector to expand the semantic features of words in short text. Clustering algorithm is an unsupervised machine learning algorithm, which divides samples into different groups based on the features. The scope of the group is not clear, because there not exist ground truth class assignments. Clustering aims to make the samples in the same group as similar as possible and the samples in different groups as different as possible.

With the cluster-based word vectors, the short text classification is carried out through Convolutional Neural Network (CNN) and Long Short-Term Memory (LSTM) neural networks respectively. Our method is evaluated on two datasets: The Chinese Medical Short Sentence (CMSS) corpus we developed and released and TREC corpus in common field. The experiments demonstrate that our method has robust superiority over competitors and sets state-of-the-art. The main contributions of this article include:

1. We propose cluster-based semantic expansion method to reduce the problems of low classification accuracy caused by feature sparseness and semantic ambiguity in medical short text classification field.
2. Based on the hierarchical aggregation clustering algorithm, we calculate the cluster embedding which represents the implicit topic information of the cluster.
3. The Chinese Medical Short Sentence (CMSS) corpus we develop and release contains 17,787 sentences that classified in three symptom severity rating, which is slightly, moderately and heavily.

The rest of the article is organized as follows. Section 2 introduces the related work about word representation and short text classification. Section 3 gives a detailed description of the overall framework of our method. Section 4 presents experimental setup, results and analysis, including the construction of Chinese medical dataset, the experiment in medical dataset and experiment in published TREC dataset. Section 5 summarizes this work and the future direction.

## 2  Related Work

The accuracy of short text classification is affected by sparseness of features since short text usually contains fewer characters. Bag-of-words (BoW) model can not be directly applied to short text representation since it expresses words using high-dimensional and sparse vectors, and ignores the order and semantic relations between words [3]. According to the distributed hypothesis, words appearing in the same context often have similar semantics. Based on the distributed hypothesis and the neural network model, Mikolov et al. [4] proposed Word2Vec model which including the Skip-Gram model and CBOW (Continuous Bag-of-Words) model. Each of the Skip-Gram and CBOW method defines a method for creating an unsupervised learning task from plain raw corpora. The CBOW model trains each word against its context, while Skip-Gram trains each the context against the word. Pennington et al. [5] proposed the global vectors for word representations (GloVe). The global log-bilinear regression model combines the advantages of the two major model families in the literature: global matrix factorization and local context window methods. The Word2Vec and GloVe models are widely used, since they consider the contextual information to obtain dense low-dimensional real-valued vectors and thereby overcome the shortcomings of BoW model.

To improve the accuracy of short text classification, many studies adopted deep neural network for the short texts classification. Facebook researchers Joulin A. et al. [6] proposed FastText model to represents sentence vectors using the mean of word vectors. Kim et al. [7] proposed the two-channel CNN using both statically and dynamically updated word embedding as input. As we all know, CNN could only capture local features, but ignore long-distance dependencies between words. Socher et al. proposed Recursive Neural Network (RNN) model for the sentiment classification [8]. It cannot obtain satisfied classification accuracy due to the vanishing gradient problem. Hochreiter S. et al. [9] employed LSTM to solve the aforementioned problem by replacing a single unit with a more complex memory unit. Zhou et al. [10] proposed C-LSTM model to represent and classify sentences. This model extracts high-dimensional phrase vectors using convolutional layer, then feeds phrases vectors into LSTM layer to get sentence vector. The model performs well since captures both local features of the phrase and the global features of sentence.

In the use of clustering, Song et al. [11] proposed the cluster-based multiple SVM classifiers to classify polarity of short product reviews. Wang et al. [12] proposed the method to classify short texts based on semantic clustering and CNN model. This method finds semantic clusters based on searching density peak, and uses n-gram to detect candidate semantic units in short text.

Combine the advantages of the aforementioned methods, we propose the cluster-based semantic expansion method to improve the performance of short text classification in medical field. We perform experiment on both CMSS and TREC data sets.

## 3 Methodology

### 3.1 Cluster-based Semantic Expansion Model

We propose a cluster-based semantic expansion model to reduce the problem of sparse features in short text. As shown in Fig. 1, we first obtain word embedding from unlabeled medical corpus through unsupervised Skip-Gram model. Then, we use hierarchical agglomerative clustering to cluster words in labeled corpus. Finally, we obtain word-cluster embedding through the concatenation of word embedding and cluster embedding.

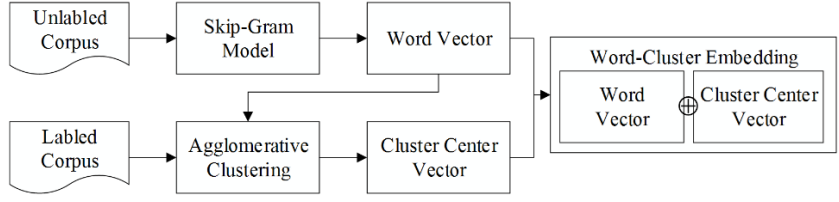

**Fig. 1.** Framework of cluster-based semantic expansion model

**Word Embedding.** Skip-Gram is used to train the context of words to obtain word embedding based on the unlabeled medical corpus. The Skip-Gram's objective function which should be maximized is presented in Eq(1):

$$J(\theta) = \frac{1}{T}\sum_{t=1}^{T}\sum_{-c \leq j \leq c, j \neq 0} \log p(w_{t+j}|w_t) \qquad (1)$$

In Eq(1), $w_t$ and $w_{t+j}$ indicate to the middle and context words separately, and $c$ points to the size of training window. Given the word $w_t$ in the middle, we can compute the log probability of predicting word $w_{t+j}$ from $-c$ to $c$ in the training window. The value of $p(w_{t+j}|w_t)$ can be calculated by Eq(2), where $v_{w_t}$ and $v_{w_{t+j}}$ represents the distribution representations of the middle and context words respectively. Through Eq(1) and Eq(2), we can finally obtain the low-dimensional dense word embedding.

$$p(w_{t+j}|w_t) = \frac{\exp(v_{w_{t+j}}^T v_{w_t})}{\sum_{w=1}^{W} \exp(v_w^T v_{w_t})} \qquad (2)$$

**Hierarchical Agglomerative Clustering.** Hierarchical Agglomerative Clustering (HAC) starts with every single sample in a single cluster, then iteratively merges two most similar clusters until there is only one cluster or preset number of clusters. HAC tends to produce smaller clusters with reasonable preset number of clusters. Therefore, it is suitable for word clustering which includes fewer words per cluster. As shown in Fig. 2, HAC produces tree structure in bottom-up direction.

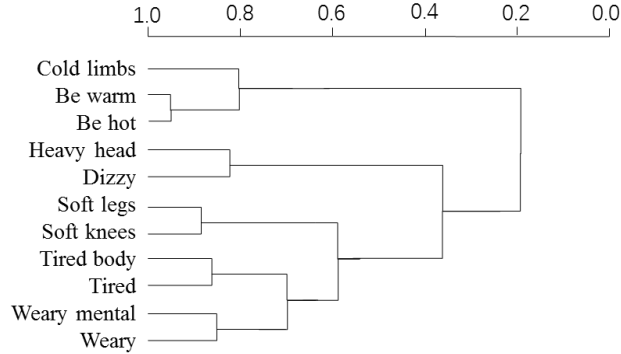

**Fig. 2.** The example of HAC in medical field

The selection of similarity function is critical. The common similarity functions include single linkage, complete linkage and average linkage. We adopt the average linkage by taking into account the sensitivity to outlier, global quality and time complexity. The other two functions are deprecated due to two reasons. 1) the single linkage calculates similarity by the distance between nearest samples in two clusters, so the cluster has good local consistency but with poor global quality. 2) the complete linkage produces compact clusters by calculating the similarity based on the distance between furthest samples in two clusters. Complete linkage is sensitive to outliers.

Given two clusters A and B, their similarity can be calculated by Eq(3):

$$sim(A,B) = \frac{\sum_{u \in A, v \in B} sim(u,v)}{size(A) * size(B)} \quad (3)$$

Here, sim(u,v) represents the similarity between the samples u and v. As shown in Eq(4), the sim(u,v) represents the Euclidian distance-based similarity:

$$sim(u,v) = \frac{1}{1+\sqrt{\sum_{j=1}^{n}(u_j-v_j)^2}} \quad (4)$$

We calculate the cluster embedding of each cluster by Eq(5). Here, m represents number of words in a cluster, while $V_i$ represents the vector of the ith word.

$$C = \frac{1}{m}\sum_{i=1}^{m} V_i \quad (5)$$

We alleviate the problem of ambiguous synonyms and feature sparseness in short text by the application of cluster embedding. The cluster embedding represent the implicit topic of all words in a cluster.

### 3.2 Short Text Classifiers

We use CNN and LSTM models to classify the short texts. Take LSTM as an example (see Fig.3), the first layer of the network is word embedding layer which transforms words into representations that capture syntactic and semantic information about the words. Each word is converted into a real-valued vector. Therefore, the input to the next layer is a sequence of real-valued vectors. We choose Word2Vec that is pre-trained on a corpus containing medical texts. Average-pooling is adopted to conduct pooling operations. The Softmax layer outputs the probabilities of short texts belonging to different categories.

The CMSS data sets we developed is classified in three symptom severity rating which is slightly, moderately and heavily. The greater the probability is, the more likely the symptom severity rating it belongs to.

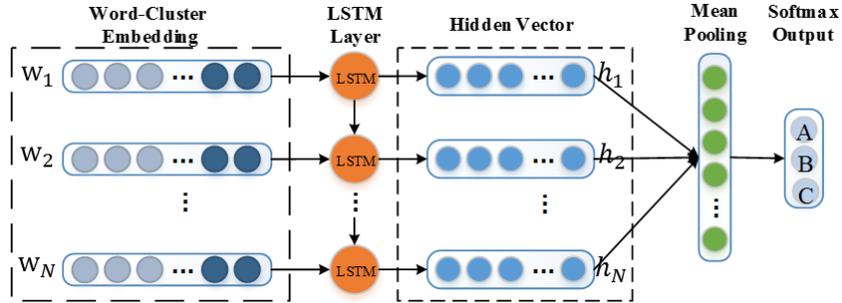

**Fig. 3.** The classifier using LSTM model

## 4 Experiment

### 4.1 Datasets

**Word Embedding of Medical Terms.** Five medical data sets and one open-domain data set is adopted in our study. As shown in Table 1, the general EMR (112,262 medical records) contains 312.7 million words, the stroke EMR (4,588 medical records) contains 1.2 million words, and the medical textbook (the 7th edition of Internal Medicine) contains 400,000 words. We obtain 490,000 word vectors using Skip-Gram model based on the aforementioned data sets.

**Table 1.** The medical corpus used to train Skip-Gram model.

| Corpus | Words(million) |
|---|---|
| Baidu Encyclopedia (medical)[1] | 37.8 |
| Hudong Encyclopedia (medical )[2] | 39.2 |
| Chinese Wikipe[3] | 44 |
| General EMR | 312.7 |
| Stroke EMR | 1.2 |
| 7th edition of Internal Medicine | 0.4 |

To solve the Out-Of-Vocabulary problem, we simply use the zero-valued vector to represent the words not appearing in the training corpus. The long term in medicine can lead to the emergence of word segmentation problems. For example, "器质性脑病综合症" could be mistakenly divided into three words, "器质性", "脑病" and "综合症". Segmentation problems may cause ambiguity and decrease the accuracy of short text classification. To this end, we construct a medical dictionary which contains 21,483 words based on ICD-10 Disease Codes and Sogou Medical Thesaurus . Jieba segmentation tool is employed for the medical corpus word segmentation with the help of the medical dictionary we construct.

---

[1] https://baike.baidu.com/science/medical
[2] http://fenlei.baike.com/%E5%8C%BB%E5%AD%A6/
[3] http://download.wikipedia.com/zhwiki/latest/zhwiki-latest-pages-articles.xml.bz2

**Chinese Medical Short Sentence.** In this paper, we focus on short text classification in the medical field in Chinese language. We obtain 2,415 short sentences from the *Guiding Principles of New Clinical Drug*, then we extend the corpus with the Chinese Wikipedia redirect dictionary that contains 640,000 synonym pairs. If a word in short sentence exists in the Wikipedia redirected vocabulary, the original word will be replaced by its synonym, and therefore a new short sentence is generated. For example, the short sentence "slightly flu" could be converted into "slightly cold" since "flu" and "cold" are synonyms.

Based on these short sentences, we construct and release a Chinese Medical Short Sentence (CMSS) corpus, to make our work more reproducible. This corpus contains a total of 17,787 sentences, among which, the sentences related to symptom severity rating slightly, moderately and heavily are 5,263, 6,072 and 6,452 respectively.

### 4.2 Implementation Details

We use word-cluster embedding as input matrix, and use CNN and LSTM model to classify short text. The CNN model is operated with two convolutional layers, two max pooling layers and one fully connected layer. Considering the training efficiency, the number of convolution kernels is set to be 64, the kernel size is set to be 5, the pooling window size is set to be 2, and the batch size is set to be 128. The LSTM is operated with one LSTM layer, one mean-pooling layer and one fully connected layer. The number of LSTM cell is set to be 300, while the batch size is set to be 128.

### 4.3 Experiment based on Medical Dataset

**Experiments based on Different Cluster Algorithms.** We employ HAC to cluster 2,718 words in CMSS data sets, and use Grid Search to find the decent number of clusters in range of 200 to 2000. To compare the performance of different clustering algorithms in short text classification, we adopt word embedding without cluster-based expansion as input matrix in different baseline models. We perform three different experiments separately using Affinity Propagation Clustering (APC), Density-Based Spatial Clustering of Applications with Noise (DBSCAN) and HAC to expand word embedding. Finally, we use CNN and LSTM model to extract features and classify short text.

As shown in Fig.4, both CNN and LSTM model using HAC-based word-cluster embedding achieve the best results. The reason may lie in: (i) the HAC clusters words in bottom-up direction, which can produce small clusters when given reasonable number of clusters. Therefore, it is suitable for word clustering which usually contains fewer words in a cluster. (ii) we use the average linkage to calculate the similarity between clusters, which is not sensitive to outliers.

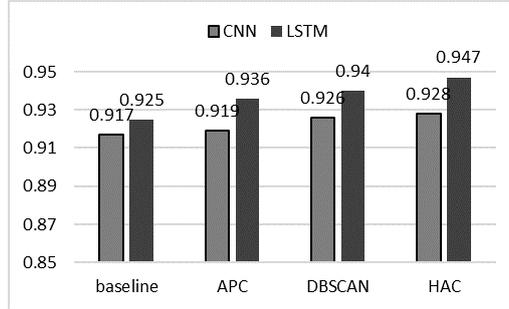

**Fig. 4.** Experimental results based on different cluster algorithms

**Experiments based on Different Classification Algorithms**. To evaluate the performance of our cluster-based word-cluster embedding, we compare our method with baselines in CMSS data sets. LibLinear [13] is an open source library for large-scale linear classification, which supports logistic regression and support vector machines. RCNN [14] uses recurrent structure to capture contextual information and max pooling layer to capture key components in text. C-LSTM [10] utilizes CNN to extract higher-level phrase representations, and employs LSTM to obtain the sentence representation.

As shown in Table 2, the HAC-CNN model achieve similar accuracy to C-LSTM, but slightly poor compared with RCNN. The HAC-LSTM model outperforms all baselines.

**Table 2.** Experimental results based on different classification algorithms

| Model | Accuracy | Reported in |
| --- | --- | --- |
| LibLinear | 0.908 | Fan et al. 2008 |
| C-LSTM | 0.929 | Zhou et al.,2015 |
| RCNN | 0.933 | Lai et al.,2015 |
| HAC-CNN | 0.928 | Our method |
| **HAC-LSTM** | **0.947** | **Our method** |

### 4.4 Experiment based on Common Dataset

To further evaluate the performance of our method, we perform experiments on public data sets: TREC. The TREC released by the UIUC Cognitive Computation Group contains 6 categories with over 6,000 labeled data. To obtain word embedding vectors, we use Google's open-source word vector trained in Google News (about 100 billion words), which contains roughly 3 million words and phrases, each of 300 dimensions.

We use HAC to cluster words and adopt cross-validation to find decent number of clusters. The experimental results show that model perform best when the number of clusters is 1000. We employ the cluster-based word-cluster embedding as input matrix, and use CNN and LSTM model to extract features and classify sentences.

9We use the following baselines: CNNs-non-static [7] uses word vectors fine-tuned during the training, and CNNs-multichannel both uses static and non-static word vectors. TFIDF + SVMs model [12] classifies the sentences using SVM classifiers based on the Term Frequency (TF) and Inverse Document Frequency (IDF) of each word in a sentence. Semantic-CNN [12] expands word embedding based on density peak clustering, and classify sentences using multi-scale CNN. SVMs [15] model classifies short text using SVM classifiers based on 60 features artificially extracted such as unigrams, bigrams, POS tags, syntax analysis, ephemera, WordNet and so on. DCNN uses the dynamic k-max pooling layer [16] to extract global features. DCNNs uses n-gram information from dependent trees [17] to capture long-range dependencies. Tree CNN [18] uses the information from the dependent tree to enrich the semantic information of word embedding.

**Table 3.** The results of short text classification on TREC data sets

| Model | Accuracy | Reported in |
| --- | --- | --- |
| SVMs | 0.95 | Silva et al.,2011 |
| TFIDF+SVMs | 0.943 | Wang et al.,2015 |
| CNNs-non-static | 0.936 | Kim,2014 |
| CNNs-multichannel | 0.922 | Kim,2014 |
| DCNN | 0.93 | Kalchbrenner et al.,2014 |
| DCNNs | 0.956 | Ma et al.,2015 |
| Tree CNN | 0.96 | Komninos et al.,2016 |
| Semantic-CNN | 0.956 | Wang et al.,2015 |
| C-LSTM | 0.946 | Zhou et al.,2015 |
| RCNN | 0.96 | Lai et al.,2015 |
| **HAC-LSTM** | **0.977** | **Our method** |

As shown in Table 4, our model outperforms the state-of-the-art on the TREC data sets. The improvement is mainly due to the cluster-based semantic expansion method, which reduces the problems of low classification accuracy caused by feature sparseness and semantic ambiguity.

## 5    Conclusion and Future Work

In this study, we propose a cluster-based semantic expansion method based on hierarchical agglomerative clustering, which effectively incorporate word embedding into cluster embedding to obtain more semantic information. Experimental results on two benchmark datasets demonstrate the superiority of our proposed method on short text classification task. We believe that the most promising avenues for future research include experimenting with methods of named entity recognition method to improve the word segmentation.